# A Survey of Recent Machine Learning Solutions for Ship Collision Avoidance and Mission Planning

**Pouria Sarhadi\*, Wasif Naeem\*\*, Nikolaos Athanasopoulos\*\***

*\* School of Physics, Engineering and Computer Science, University of Hertfordshire, Hatfield, UK, (e-mail:*
*p.sarhadi@herts.ac.uk)*
*\*\* School of Electronics, Electrical Engineering and Computer Science,*
*Queen's University Belfast, Belfast, UK, (e-mails: w.naeem@qub.ac.uk, n.athanasopoulos@qub.ac.uk)*

Abstract: Machine Learning (ML) techniques have gained significant traction as a means of improving the autonomy of marine vehicles over the last few years. This article surveys the recent ML approaches utilised for ship collision avoidance (COLAV) and mission planning. Following an overview of the ever-expanding ML exploitation for maritime vehicles, key topics in the mission planning of ships are outlined. Notable papers with direct and indirect applications to the COLAV subject are technically reviewed and compared. Critiques, challenges, and future directions are also identified. The outcome clearly demonstrates the thriving research in this field, even though commercial marine ships incorporating machine intelligence able to perform autonomously under all operating conditions are still a long way off.

## 1. INTRODUCTION

The use of Artificial Intelligence (AI) and Machine Learning (ML) has gained momentum for a variety of challenges around autonomous vehicles and related fields (Ma et al., 2020, Aradi, 2020, Kuutti et al, 2021); Autonomous and electric vehicles are taking the lead after smartphones as the main outlets to demonstrate and promote digital technology. Maritime vehicles are widely accepted as efficient transportation systems and are therefore a direct beneficiary of this research and development. Some key studies such as Campbell et al., (2012) have suggested the importance of AI exploitation in reducing human errors and in preventing collision of maritime surface vehicles. Further, the significant number of recent industrial projects pitching autonomous functions in ship mission planning and control emphasises the importance of this topic. The projects Mayflower (2022), Yara (2021), L3HARRIS (2021), Artemis (2020), Cetus (2020), and MAXCMAS (2018) are some of the most recent examples of prominent industry-led research aiming at the development of autonomous and high-tech vessels for maritime applications. A simple literature search reveals a significant increase in research publications dealing with ML applications to autonomous ships. Classical approaches in path planning and collision avoidance of ships are continuously investigated (Tam et al., 2009, Campbell et al., 2012, Huang et al., 2020, Zhang, et al., 2021a, Vagale et al., 2021a and 2021b, Li et al., 2021a, Ozturk, et al., 2022); Nevertheless, this survey is concerned with a different aspect, namely, the recent advances in ML techniques for mission planning and collision avoidance of maritime vessels.

The applications of ML techniques to other types of (ground, underwater and aerospace) autonomous vehicles have been reported in comprehensive research articles. Recent studies include motion planning and control of autonomous cars (Aradi, 2020, Kuutti, et al., 2021, and Kiran et al., 2021), intelligent transportation systems (Haydari et al., 2020), robotics (Kroemer, et al., 2021, Ibarz et al., 2021, and Sun et al., 2021), Unmanned Aerial vehicles (UAVs) (Fraga-Lamas, et al., 2019), Autonomous Underwater Vehicles (AUVs) (Hadi et al., 2021), and spacecraft control system design (Shirobokov et al., 2021). However, despite the considerable technological and commercial relevance of autonomous ships, this topic is less explored. In this paper, recent ML solutions for ship collision avoidance and mission planning are technically surveyed. It should be noted that the focus of this paper is on the mission planning and COLAV applications. Besides, the references provided in this article could be considered as a bibliography for the recent advances in autonomous ship design. There are several terms in the literature used to describe these vehicles, including Unmanned Surface Vehicles (USV), Autonomous Surface Vehicles (ASV), Maritime autonomous surface ships (MASS), nonetheless, USV is preferred in this paper. Following a brief introduction to ML and mission planning for USVs, state-of-the-art advancements and research in the field are presented. An attempt is made to categorise and compare existing solutions and to draw out their shortcomings and envision the potential next steps.

The rest of this paper is organised in the following form. Section 2 outlines the application of ML techniques to USVs. In Section 3, areas to be addressed in the mission planning of an autonomous vessel are introduced. Sections 4 and 5 review and compare the existing research with direct and indirect applications to the planning problem, respectively. In Section 6, achievements, challenges, and future directions in this research topic are identified based on the surveyed papers. Finally, Section 7 concludes the paper.

## 2. BACKGROUND: MACHINE LEARNING AND ITS APPLICATION TO USVs

Even though ML and AI are not new topics in engineering or data science, recent progress in Deep Learning (DL) techniques could enable AI usage for complex autonomous functions (Goodfellow et al., 2016, Li, 2017, Sutton and Barton 2018). Deep Reinforcement Learning (DRL), due to its affinity with control theory and learning ability via feedback from a reward function, occupies a prominent position in intelligent mission planning and control applications (Kiran et al., 2021). Terms such as agent, environment, and action in RL are substituted by controller, controlled system (or plant), and control signal respectively (Sutton and Barton 2018). In this study, it was found that DRL has been dominant in the ship mission planning topic. However, other AI solutions are also proposed in the literature. In fact, AI, ML, DL, RL algorithms used in control systems and autonomous vehicles share interconnections. Fig. 1. depicts the intersections amongst those topics in a Venn diagram as an extension of the one in (Goodfellow et al., 2016) to consider control systems and autonomous vehicle algorithms.

The present study considers direct and indirect ML applications for the mission planning of USVs in the past five years. The direct applications consider those algorithms that have been exploited for planning and collision avoidance purposes (Section 4). The second category of indirect applications encompasses techniques which are not used for collision avoidance but have the potential to be applied in other relevant topics such as risk assessment and global planning (Section 5). Those definitions and use-cases will be discussed in more detail within Sections 3-5.

It is worth noting that there exist some very interesting papers that consider topics in control and perception within USVs, however, they are not the focus of this survey. For instance, Martinsen et. al., (2022) have implemented RL-based nonlinear model predictive control for trajectory tracking of fully actuated vessels. A Deep Q-Network (DQN) is exploited for the system identification part of the controller. The proposed technique was tested on the ReVolt USV during square sides dynamic positioning. In another example, Du et al., (2022) developed a Lyapunov boundary deep deterministic policy gradient (DDPG) for a USV for vessel tracking and interception tasks. In the proposed strategy, a combination of line of sight (LOS) proportional guidance and neuron adaptive learning control were employed in the own ship (OS) to pursue a target ship (TS) and intercept it. The technique was implemented in a Gazebo-based virtual reality simulator. Other cases are also reported for formation control of USVs in Wang et al., (2021), auto-docking in Gjærum et al., (2021a, b), boat autopilot in Cui et al., (2019) and (2021), and path following control in Gonzalez-Garcia (2020). Nevertheless, in this article, mission planning (described in Section 3) applications are investigated, and the reviewed papers are categorised into the aforementioned direct and indirect groups. It is important to note that this review is concentrated on key relevant papers that published since 2018. Due to space limitations, detailed mathematical descriptions of the algorithms are out of the scope of this paper. The interested reader can avail of the references provided.

Based on a thorough search of this topic, there appears to be a significant increase in published research in this area. Fig. 2 represents the number of published articles in recent years (until June 2022). Based on this figure, an exponential increase is anticipated in the upcoming years considering the publication time of this manuscript. It should also be mentioned that scientific databases such as Google Scholar, Scopus, and the main publishers such as IEEE, ScienceDirect, etc. are utilised for this research, due to different alternative terminologies and titles used to denote these systems (USV, ASV, MASS, etc.). Fig. 3 illustrates a word cloud for the keywords utilised in the reviewed papers which reveals the diversity of keywords used in this topic. Some of the key words in Fig. 3 will be further discussed in the following sections.

## 3. MISSION PLANNING PROBLEM Of MARITIME VESSELS

In general, the planning or mission planning problem is related to generating feasible paths or trajectories to be tracked by a vessel. The mission may be pre-defined by a human operator or modified during the journey, either remotely or by the onboard crew. To this end, several key areas, specified and grouped in Fig 4, should be considered. Those topics are outlined in this section, and they are used as the foundation of the comparison between the reviewed ML techniques.

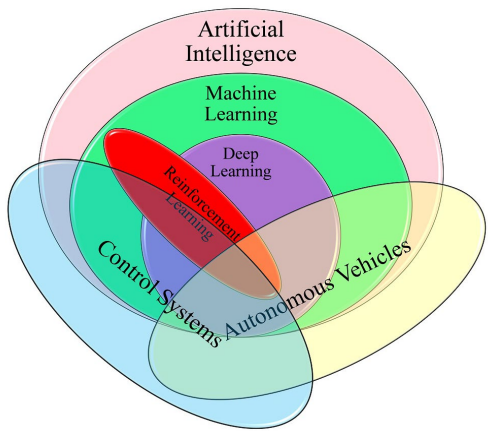

Fig. 1. A Venn diagram illustrating the relationship between AI, ML, DL, RL, algorithms of control systems and autonomous vehicle algorithms.

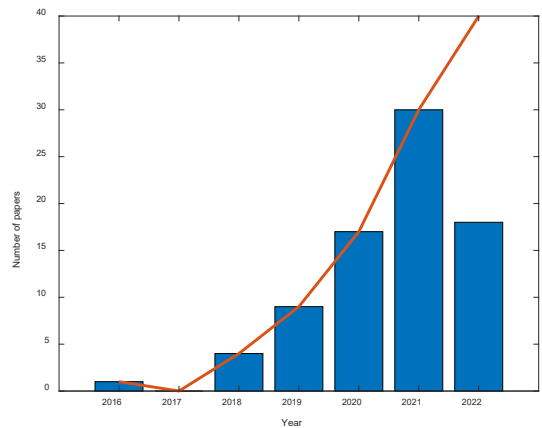

Fig. 2. Number of yearly publications on this survey's topic, predicting exponential growth in the upcoming years.

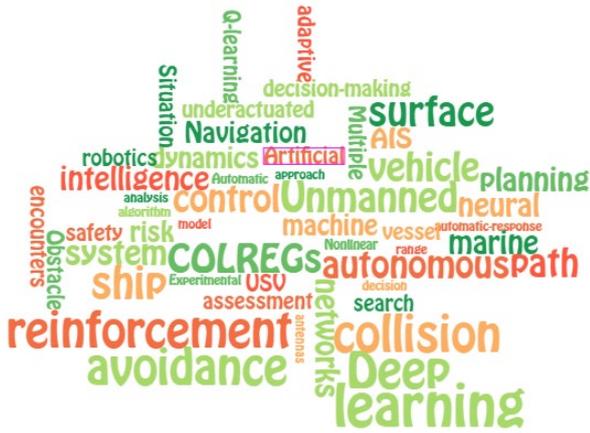

Fig. 3. Keyword cloud in relevant papers of this survey.

### 3.1 Global Planning

Global planning algorithms generate a feasible set of waypoints for a mission. Different aspects such as optimality (in terms of path, time, fuel consumption etc.), adhering to maritime rules, feasibility based on an updated map, to name a few are considered at this level. As an instance, Fig. 5 shows how successive waypoints (WP1-20) are generated to traverse between Oslo and Trondheim (Zhang et al., 2021a).

### 3.2 Local Planning

Local planners are algorithms that plan the motion in terms of detailed paths or trajectories to be tracked by the vessel and define how to traverse between global waypoints. Note the difference between path and trajectory lies is in the inclusion of time in trajectory planning, whereas, in path planning, the time to reach a certain point is less important. In practice, a straight line is the shortest path between two waypoints, and algorithms are designed to minimise the Cross Tracking Error (CTE) to adhere to the shortest path (Fossen, 2011). This task is sometimes called path or trajectory planning. Nonetheless, traversing a straight-line path between any two waypoints is not always possible due to the presence of obstacles and environmental disturbances. Regardless, collision avoidance could also be needed even if LOS is maintained.

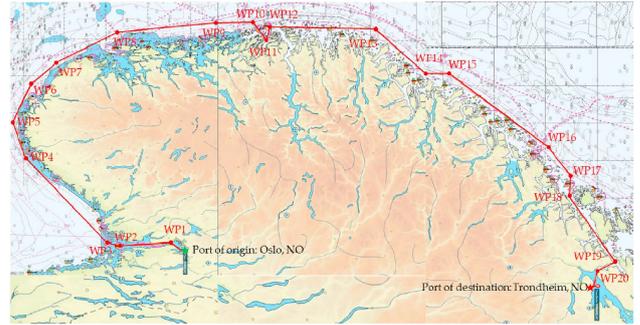

Fig. 5. Global waypoints planned ahead for a maritime journey from Oslo to Trondheim (from Zhang, et al., 2021a).

### 3.3 Collision Avoidance (COLAV)

The goal of COLAV is to modify the planned path or trajectory in such a way that prevents any collision with an obstacle. COLAV is sometimes viewed as a sub-task of local planning, but here it is considered an essential component of mission planning. Those obstacles could be static including isles, buoys, maritime infrastructures, etc. or dynamic such as other vessels, drifting objects, animals, etc. An illustration of path following, and COLAV is depicted in Fig. 6 where OS is the own ship, $\psi$ is the desired heading angle and $WP(i) = [X_{wp(i)}, Y_{wp(i)}]$ defines waypoints, $i$ being the index. As can be seen from Fig. 6, the planner should meet the waypoint tracking objective whilst avoiding collision with the obstacles simultaneously.

### 3.4 COLREGs

In maritime transportation, power-driven vessels approaching each other should adhere to certain rules. These rules are defined by the International Maritime Organisation IMO (1972) and named Convention on the International

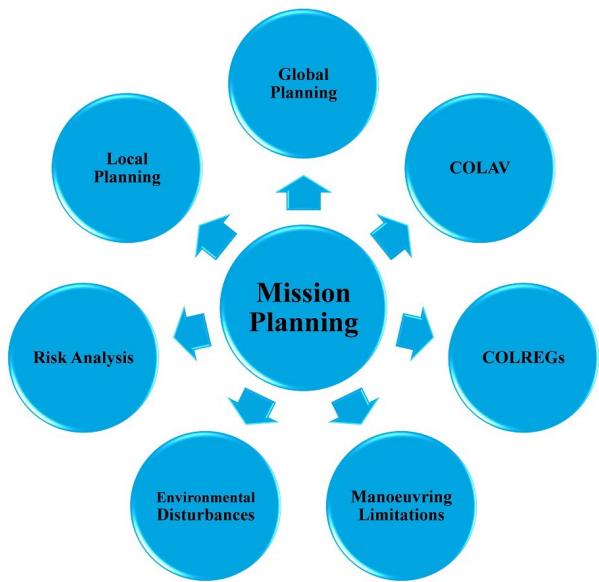

Fig. 4. Different subjects to be considered in the mission planning of a ship.

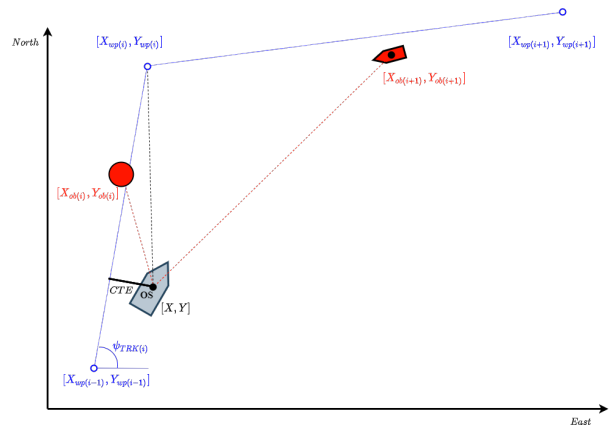

Fig. 6. Path planning and collision avoidance in local planning.

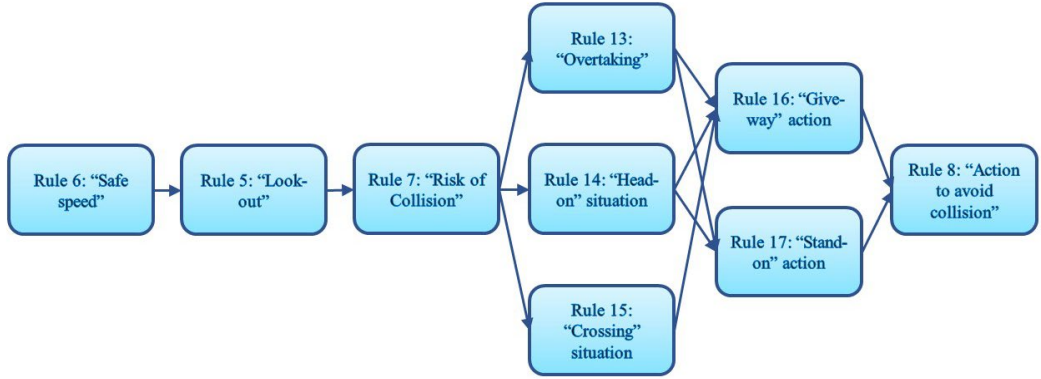

Fig 7.   The process for COLREGs based decision making

Regulations for Preventing Collisions at Sea (COLREGs). Fig. 7, adopted from Namgung and Kimg, (2021) depicts a typical process to apply COLREGs. Rules 5-8 and 13-17 can be directly exploited for the mission planning of ships. Based on Fig. 7, and COLREGs rules, each ship should travel at a safe speed and monitor for close encounters. In case of a risk involving other ships in the vicinity, one or more of the three fundamental encounter situations including Overtaking, Head-on, and Crossing situations could be identified. Appropriate action in the form of give-way or stand-on should be undertaken to avoid any collision. Translating these rules to the algorithms is challenging as the rules were originally written for human consumption. Learning these rules should be incorporated into the ML approaches.

### 3.5 Risk Assessment

One of the crucial factors to be considered in mission planning is risk assessment. Risk is usually defined in terms of an index (i.e., Collision Risk Index- CRI) that shows how likely a hazardous event such as a collision could happen. Various approaches are proposed to calculate the collision risk index (Huang et al., 2020a, Pietrzykowski, and Wielgos, 2021), however, most of them are based on the Closest Point of Approach (CPA) analysis. The CPA determines how close two ships would come to each other if they both continue to move at the same speed and direction. Distance to CPA (DCPA) and Time to CPA (TCPA) form the basis of most risk assessment techniques (Huang et al., 2020a). The computation of those parameters is based on the relative velocity vector ($V_{rel}$) and its relative angle ($\alpha$) between the OS and TS, as shown in Fig. 8. For further information about the calculation procedure of DCPA and TCPA, one can refer to Sarhadi et al., (2022).

### 3.6 Manoeuvring Constraints

Ship manoeuvring constraints involving non-holonomic behaviour, underactuated dynamics, system response lag, control limitations, etc. are important issues that should be carefully weighed in local planning. Those items are considered in the ship modelling and dynamics prediction approach. In the literature, diverse models are utilised to model and predict the behaviour of OS and TS (Huang et al., 2020a). In addition, the closed-loop control behaviour is another item to be included in planning to generate pragmatic paths. A highly precise model can result in a more practical machine learning algorithm. Hence, a proper planning approach should consider the aforementioned manoeuvring limitations. As mentioned, research in the field of risk assessment is vibrant and the interested reader is referred to Chen et al., (2019a), Huang et al., (2020a, b), Du et al., (2021) for further information about risk analysis in ship manoeuvre.

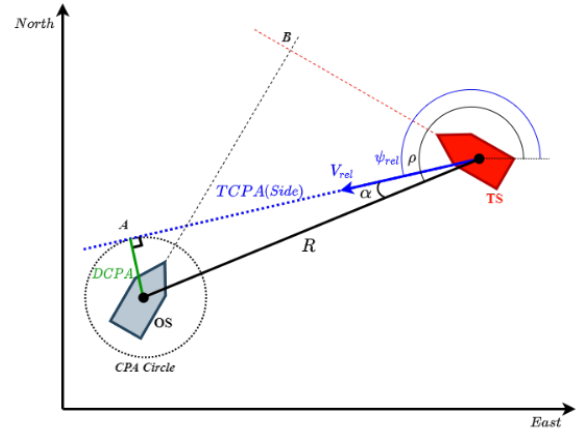

Fig. 8. The CPA illustration between OS and TS to calculate TCPA

### 3.7 Environmental Disturbances

To improve the precision of vessel models, environmental disturbances should be considered. To this purpose, high-fidelity ship models consider the effect of waves, winds, and currents in their models (Fossen, 2011). It is clear that due to the impact of the environmental disturbances in vessel motion, these phenomena should be modelled in the learning procedure, or at least tested after the learning phase in ML-based mission planning and COLAV design.

Other topics can be considered in the algorithm design however, it is believed that 2.1-2.7 form the touchstone for comparison between surveyed ML approaches in this paper.

## 4. ML WITH DIRECT APPLICATIONS TO PATH PLANNING AND COLLISION AVOIDANCE

In this section, ML applications directly utilised for (local) path planning and collision avoidance for autonomous vessels are discussed.

Zhao and Roh (2019) proposed DRL for the collision avoidance of multiple USVs. In that study, two layers of fully connected (FC) multilayer perceptron neural networks are employed with a proximal policy optimization (PPO) learning algorithm. The vessels are modelled by 3DOF equations and disturbances were not considered. The output of the ML algorithm is the rudder angle for which amplitude and rate constraints are also considered for the actuation system. The reward function comprises of reaching the goal, heading error, cross tracking error, drift, collision to obstacles, and COLREGs. Simulation results in crossing, head-on and overtaking scenarios for multiple ships are presented. As an additional parameter, the control signal time history is revealed, which shows aggressive behaviour. This study is an expansion of research in Zhao et al., (2019). In Cheng and Zhang (2018), a concise DRL obstacle avoidance algorithm is developed for underactuated unmanned marine vessels. The yaw moment and the surge propulsion forces are considered as the action spaces for the ML algorithm. To overcome the discrete nature of DQN, the actions are selected from a space of one unit increase or decrease, or the previous identical value. The defined reward is based on the distance to the goal and the obstacle, drift from the straight path, and speed at the goal position. Some preliminary simulations are carried out in a limited space with several static obstacles. Risk and COLREGs topics are ignored in the planning, and excessive input efforts were visible in exhibited simulations.

Xu et al., (2019) developed a DDPG for the same problem. For data generation, a dynamic model was utilised, whilst thrust and turning moments are the action spaces. Distance to target, distance to obstacles, maximum lateral velocity to prevent drift, and speed reduction near the goal are reward function indices. Risk and COLREGs were not taken into account in planning, and simulations were exhibited for static obstacles. In Xie et al., (2019), a Model Predictive Control (MPC) via an Improved Q-learning Beetle Swarm Antenna Search (I-Q-BSAS) and ANN (to estimate an inverse model for the optimal policy approximation) are merged for multi-ship collision avoidance. A combination of CRI, LOS tracking error and rudder optimisation was defined for the MPC optimisation problem which produces the rudder angle as the output. COLREGs and CPA-based risk analysis were also incorporated in planning. The proposed approach was tested on the KVLCC2 (Xie et al., 2019) ship model in small and large-angle crossings, overtaking and head-on scenarios. In addition, the performance of I-Q-BSAS was compared against various optimisation techniques showing better results. Moreover, in Xie et al. (2020), a model-free RL-based multi-ship collision avoidance algorithm was developed which combined an asynchronous advantage actor-critic (A3C) algorithm, a long short-term memory neural network (LSTM) and Q-learning. The LSTM part is exploited to accelerate the model-free A3C learning by adaptive Q-learning decisions. The reward function embraces a CRI, control action and LOS tracking error and generates the required rudder angle for path following and COLAV. It should be noted, in this research, COLREGs were not considered.

Meyer et al., (2020a) conducted a comparative study between state-of-the-art DRL techniques i.e., PPO, DDPG, Actor-Critic using Kronecker-Factored Trust Region (ACKTR), and Asynchronous Advantage Actor-Critic (A3C) techniques. The comparison was carried out for the collision avoidance of a 3DOF model of Cybership II in OpenAI gym Python toolkit. Better performance for PPO has been reported for diverse path following and collision avoidance scenarios. Nevertheless, the authors outlined the challenge of the application of a massive number of NN parameters to safety-critical systems. The same researchers presented a PPO-based COLREGs-compliant collision avoidance for Cybership II USV model in Meyer et al., (2020b). A reward function that considers path following (based on CTE and speed) and collision avoidance (separately for static and dynamic obstacles) were proposed. The performance of this approach was tested on map-based static obstacles and some COLREGs situations. Interesting simulation scenarios were illustrated since some of them were carried out in real map situations. However, the environmental disturbances were ignored which is a shortcoming. In Meyer et al., (2022) the former research is improved to include a CPA-based risk-analysis in COLAV decision making. In Liu and Jin (2020), a combination of deep Q-network (DQN), double DQN, and duelling DQN has been proposed for ship collision avoidance. Seven discrete speed and angular rate choices were separately considered as action spaces. A positive reward for reaching the goal and negative commands for hitting an obstacle or stopping were considered. Nevertheless, risk and COLREGs are not taken into account in decision making. For simulation, points in a 2D gaming environment were deployed.

In Chun et al., (2021), the research in Zhao and Roh (2019) has been comprehensively extended to incorporate the risk analysis in the planning phase. The CPA and ship domain are simultaneously utilised for risk assessment whereas the PPO is the main learning algorithm. The defined reward function is based on two categories: 1) path following to include reaching the goal, CTE and check points (waypoints); 2) COLAV, which includes the risk-based collision avoidance and COLREGs. The control actions are selected from a discrete vector containing zero, minimum and maximum rudder rates. Although the proposed technique was simulated in a large area, it was restricted to one similar azimuth. The proposed technique is compared to the conventional A* algorithm and a better performance is claimed.

Sawada et al., (2021), have proposed an idea to employ LSTM in PPO-based DRL to generate continuous COLAV actions. The inputs to the network are fed from a grid sensor (a virtual sensor to perceive the OS and TS locations), waypoints, and own ship's information. Obstacle zone by target (OZT) is established to consider the risk of collision. A reward function based on the distance to waypoints, move to starboard, yaw stability, COLAV and arriving at the target point is developed. The outputs of the ML algorithm are heading command, and rudder angle. COLREGs are incorporated in only move to starboard preference. Good practice in this article is to use 22 standard encountering test scenarios called the Imazu problem. The same benchmark problem is exploited in Zhai et al., (2022) to assess the developed COLAV technique. In the future, elaborated benchmark scenarios will be required to compare the safety and performance of emerging proposed algorithms (either ML or classical).

**Table 1. Comparison among selected research on ML-based path planning and COLAV for USVs**

| References | Applications | ML Approach | Vehicle model for learning | ML output | Risk | Performance indices | Testing scenarios |
|---|---|---|---|---|---|---|---|
| Zhao and Roh (2019) | COLAV for multiple USVs | Two layers FC DNN PPO | Dynamic 3DOF simulation | Rudder angle | No | Reaching the goal, heading error, cross error, drift, collision, COLREGs | Crossing, heading, overtaking |
| Xie et al., (2019) | Model predictive ship collision avoidance | MPC I-Q-BSAS ANN | Dynamic 3DOF model of KVLCC2 ship | Rudder angle | Yes (CPA) | CRI, LOS tracking error and rudder optimisation | Small- and large-angle crossing, overtaking and heading on |
| Woo and Kim (2020) | COLAV for USVs | DQN CNN SMDP VO | Identified model of WAM-V USV and the real USV | Course and velocity | Yes (CPA) | Reward for path following (CTE), collision avoidance (TCPA and course), and COLREGs | Crossing, heading, overtaking |
| Chun et al., (2021) | Path planning and COLAV using ML considering the risk and | DNN PPO | Dynamic 3DOF model (3DoF) with wind and Current | Rudder angle | Yes (Ship domain+ CPA) | Reaching goal, cross error, check point (waypoints), risk-based collision avoidance, COLREG | Crossing, heading, overtaking |
| Meyer, et al., (2021) | COLREGs-compliant COLAV | PPO | Nonlinear model of Cybership II | Surge force and yaw torque | No | path following (CTE, and speed) and COLAV (static and dynamic based on COLREGs) | Map-based static obstacles and COLREGs |
| Li at al., (2021b) | COLREGs based collision avoidance | APF-DQN | Not considered | Nine discrete heading commands | No | Reward based on the APFs attractive and repulsive forces | Static and dynamic obstacles in small dimension |
| Swada, et al., (2021) | COLAV via a novel risk index | DRL with LSTM | Nomoto model | Heading command, rudder angle | YES (OZT) | Reward based on distance to waypoint, move to starboard, yaw stability collision and arriving | Up to three ships in COLREGS and Imazu problem |
| Xu, et al., (2022a) | USV motion planning and COLAV | Three layers FC DNN (DQN) DDPG | Dynamic 3DOF model | Trust, rudder | Yes (CPA) | distance to target, heading error, collision, COLREGs, reaching to goal, speed | heading on, crossing, overtaking in a simulator |
| Zhou, et al., (2022) | USV collision avoidance | DQN MDQN MDDPG | Dynamic 3DOF model | Rudder | No | Arriving in target and crash | heading on, crossing, overtaking in small dimension |
| Zhai et al., (2022) | Only autonomous COLAV | DDQN | Nomoto model | Course alteration | Yes (CPA) | Reward based on rudder activity, course change, path deviation, risk and COLREGs | Up to three ships in COLREGS and Imazu problem |
| Heiberg, et al., (2022) | Path following and COLAV | PPO | Dynamic 3DOF model of CyberShip II | Surge force and yaw torque | Yes (CPA) | path following (CTE, and speed) and COLAV (static and dynamic based on COLREGS and Risk) | COLREGs scenarios and static obstacles |

Zhai et al., (2022) have adopted the discrete Double DQN (DDQN) to solve the COLAV-only problem. It is assumed that the waypoint following is in-operation and only heading alterations are required to avoid any collision. Therefore, the outputs of the ML are thirteen levels of heading modification. A reward function based on rudder activity, course change, path deviation, risk and COLREGs is developed. The Nomoto model (Fossen, 2011) is utilised to train the ML. As mentioned, the performance of the proposed technique is verified in Imazu problem tests.

Fan et al., (2022) have also used DQN to generate COLAV commands. The discrete commands are in terms of rudder rate modifications. Rewards are allocated in two sets: i) final rewards, i.e., arriving at the destination, and COLREGs, ii) sample rewards, i.e., tracking and distance to the course are utilised. The Norrbin model (Fossen, 2011) of Lan Xin USV is employed to train the ML. Results in some COLREGs scenarios with one and multiple TS are presented to demonstrate the performance of the proposed algorithm and the implementation is considered as future work.

In Zhou et al., (2022) an improved DQN in terms of a Modified Deep Q Network (MDQN) and a Modified DDPG or MDDPG were developed for the collision avoidance of USVs. The memory pool, success pool, and target network were modified to smooth the training process. A relatively simple reward function is employed that only considers the arriving at the target and the obstacle avoidance objectives. The proposed schemes were compared for some COLREGs scenarios. Superior performance for MDDPG is reported, however, the considered test dimension is small (<100m). Woo and Kim (2022) proposed DRL and a semi-Markov decision process (SMDP) for USVs' COLAV. A DQN based on convolutional neural networks (CNN) and fully connected layers are developed to decide between path following or COLAV modes. Then, the velocity obstacles (VO) method is utilised for generating the necessary manoeuvre. For DRL training, a reward comprising path following (inverse of CTE), collision avoidance (including TCPA and vessel course), and COLREGs were introduced. Simulation and experimental results on the WAM-V USV platform are presented. The need for hundreds of hours of training and the adverse effect of modelling uncertainty in simulation to experiments are nominated as the main challenges of this approach.

Duelling DQN prioritized replay (Duelling-DQNPR) is proposed in Gao et al., (2022) for the mission planning of a USV with AIS data learning. The proposed approach involves embedding a vehicle model in a real map and replaying AIS traffic data to learn mission planning tasks. Using both static and dynamic factors, a reward function is developed. The state space is a vector of OS location (outputs of a 3DOF dynamics model) and TS information from the AIS data replay. No risk index or COLREGs are considered based on the smaller size and greater manoeuvrability of the USV. The action space is selected between 11 discrete rudder levels. As another notable research, Xu et al., (2022a) incorporated DRL into the COLREGs-compliant path planning and dynamic collision avoidance of ASVs. In this research, DDPG has been applied for generating thrust and rudder inputs to control the vessel. Saturations for the control inputs were also taken into account. The reward function is based on the distance to the target, heading error, collision, COLREGs, reaching the goal, and the speed. The ML algorithms have incorporated CPA-based risk in collision avoidance. Simulations are carried out in a visualised environment in different COLREGs situations with multiple target ships. A similar study to Xu et al., (2022a) is presented in Xu et al., (2022b). The literature is not limited to the above articles and similar interesting research with a broader set of objectives can be found in an interested reader may refer to: Shen et al., (2019), Zhou at al., (2019), Chen et al., (2019b), Amendola et al., (2020), Guo et al., (2020), Wu et al., (2020), Luis et al., (2021), Xu et al., (2020), Wang et al., (2021), Zhang et al., (2021a), Chen et al., (2021a), Abede, et al., (2022).

Table 1 outlines a comparison between some of the notable results in this field. It is noteworthy that all approaches have incorporated some COLREGs in the planning. Based on this table, all articles deployed DRL-based ML approaches. Selection of rudder as the main control signal and preferring constant speed are dominant in action space configurations. Diverse configurations for the reward function are developed that could be insightful. However, in the case of using a reward function with several weighted indices, a challenge could be defining the rewards' weights which may result in supervised learning thus requiring expert human knowledge.

# 5. ML WITH INDIRECT APPLICATIONS TO MISSION PLANNING AND COLLISION AVOIDANCE

This section investigates ML algorithms found in the literature that could be potentially employed in miscellaneous mission planning subtopics of autonomous ships as outlined in Section 2. These include methods that have been proposed for global planning, collision analysis, ship trajectory prediction, etc. For instance, in Kim and Lee (2018) a deep learning framework called Ship Traffic Extraction Network (STENet) is proposed for medium- and long-term traffic prediction in a specific maritime area. For this purpose, a combination of CNN and FCNN was developed that predicted the number of ships in the caution area using ship length, destination, channel type, Pilot Onboard (POB) and Caution Area Estimated Time of Arrival (CAETA) from the AIS data, and ship movement data. A comparison between Dead Reckoning (DR), Support Vector Regression (SVR), Very Deep Convolutional Networks (VGGNet) models in ship traffic prediction is conducted with Mean Absolute Error (MAE) and the standard deviation as the basis for comparing their performances. Subsequently, better performance for STENet in both medium- and long-term predictions is asserted.

Lei et al., (2019 and 2021) have modelled the human operators' behaviour in encountering near-collision scenarios via AIS data. The proposed approach is based on two parts: conflict detection (clustering) and collision avoidance behaviour learning via Generative Adversarial Networks (GAN) with a long short-term memory (LSTM) based encoder-decoder architecture. In a series of papers, Murray and Perera (2021a, b, c) have proposed an AIS-based deep learning framework for ship behaviour prediction and proactive collision avoidance. Hierarchical Density-Based Spatial Clustering of Applications with Noise (HDBSCAN) and Deep Recurrent Neural Networks (RNN) was proposed in Murray and Perera (2021a) to cluster, learn, and anticipate the ship behaviour using AIS data. Predictions based on the real motion of a ship with a 300 m mean squared error in 30 min was exhibited. In Murray and Perera (2021c), a proposal for proactive reaction considering COLREGs and CPA analysis has been proposed. Further investigation is required to verify the fidelity of the proposed scheme. Monitoring ship safety in extreme weather events and developing contextually aware ship domains via ML algorithms are respectively explored in Rawson et al., (2021) and Rawson and Brito (2021). Rawson et al., (2021) have recommended Support Vector Machines (SVM) over some ML algorithms to quantify the relative likelihood of an incident during the US Atlantic hurricane season. For this purpose, a comparison amongst SVM, Logistic Regression (LR), Random Forest (RF), Extreme Gradient Boosting (XGBoost), SVMs optimised using Stochastic Gradient Descent (SGD) and Multi-layer Perceptron (MLP) algorithms was carried out. The weather-related risks are considered in seven areas including wind speed and height, vessel category, length, flag, age, distance from shore, and incident data. Finally, a case study on the Hurricane Matthew (October 2016) demonstrated the ability to predict the accident via a likelihood score. In Rawson and Brito (2021) developing risk models via contextually aware ship domains is studied. The RF machine learning algorithm is employed for data mining of big vessel traffic datasets to identify the encounter characteristics across various situations and to predict the critical passing distance between vessels. The developed ship domain is dependent on the ship speed, size, encounter type, weather, waterway characteristics and is trained on realistic ship encounter data. The potential advantages of this method for estimating the likelihood of a collision in a crowded maritime area are discussed.

In one of the most comprehensive articles in this category, Namgung, and Kim (2021) proposed an Adaptive Neuro-Fuzzy Inference System (ANFIS) as a collision risk inference system that incorporates COLREGs and risk into the algorithm. Parameters including DCPA, TCPA, the variance of the compass bearing degree (VCD), the relative distance between own and target ship are utilised to compute a CRI using ANFIS. Risk inference for near-collision encounters based on real AIS data in a predefined area is presented. As an extension, the same authors have used a density-based spatial clustering of applications with noise (DBSCAN), a fuzzy inference system based on a near-collision (FIS-NC) and long short-term memory (LSTM) to draw out regional collision risk and assist the vessel traffic service operator. The focus is near-collision situations where two ships' domains overlap. Abede et al., (2021) have employed the Dempster-Shafer (DS) theory to estimate a collision risk index based on AIS data. The developed risk index contains vessel speed, relative speed and bearing as well as TCPA and DCPA. Gradient boosting regression (GBR) is deemed the best ML technique to enhance the DS evidence theory. Zhao et al., (2022) have proposed ship trajectory prediction via an ensemble ML algorithm to remove outliers in the raw AIS data to predict ship trajectories. For this purpose, publicly accessible databases and merchant websites have been utilised. Empirical mode decomposition (EMD) is used to suppress the AIS data outliers (for data denoising) and an ANN is utilised for the trajectory prediction. A comparison between LSTM, vondrak+ANN and wavelet+ANN in terms of prediction error is presented. Trajectory predictions for three typical ship types (i.e., container ship, cargo ship and passenger vessel) are presented in this paper. Jeon-Seok et al., (2022) have proposed a framework to generate maritime traffic routes using statistical density analysis. Hausdorff–distance, Douglas–Peucker, and DBSCAN algorithms are subsequently deployed to locate the waypoints and to create the connecting routes. The outcome is in the form of planned routes for autonomous surface ships. Some other intriguing research can be found in Ozturk et al., (2019), Shi and Liu (2020), Gao and Shi (2020a, b), Chen et al., (2021b), Park and Jeong (2021), Ivanov et al., (2021).

**Table 2. Comparison between ML-based algorithms with various applications of planning**

| Reference | Application(s) | ML approach | Risk inclusion | COLREGS |
|---|---|---|---|---|
| Kim and Lee (2018) | Medium- and long-term traffic prediction | CNN, FCNN | No | No |
| Murray and Perera (2021a, b) | Ship behaviour prediction and proactive avoidance | HDBSCAN and RNN | No | Yes |
| Lei et al., (2019) and (2021) | Prediction of maritime collision avoidance behaviour considering risk and COLREGs | Clustering, GAN, and LSTM | Yes (CPA) | Yes |
| Rawson et al., (2021) | Monitoring ship safety in extreme weather events | SVM | Yes (weather-related indices) | No |
| Rawson and Brito (2021) | Developing risk models via contextually aware ship domains | RF | Yes (ship domain) | No |
| Namgung, and Kim (2021) | Collision risk inference system | ANFIS | Yes (ship domain+CPA) | Yes |
| Namgung, and Kim (2021) | To extract near collision situations | DBSCAN, FIS-NC, LSTM | Yes (ship domain+CPA) | No |
| Abede et al., (2021) | Collision risk index estimation | Dempster-Shafer theory | Yes (CPA) | Yes |
| Jeong-Seok et al., (2022) | Global Planning (route planning for autonomous ships) | Hausdorff–distance, Douglas–Peucker, and DBSCAN | No | No |
| Zhao, et al., (2022) | Ship trajectory prediction | EMD, ANN | No | No |

In Table 2, a summary of ML algorithms not in direct relation to local planning and COLAV is presented. As could be observed from Table 2, various learning techniques are utilised and unlike local planning solutions discussed in Section 4.1, DRL is not dominant. In this application, clustering and pattern recognition techniques are more common. Nevertheless, the proposed approaches discussed in this section have potential for applications such as global route planning and risk assessment.

## 6. ACHIEVEMENTS, CHALLENGES AND FUTURE DIRECTIONS

According to the surveyed research, in recent years, a substantial effort has been dedicated to exploit advanced ML techniques to the ship mission planning and collision avoidance problems. Despite the progress achieved, this topic remains in its infancy with a long voyage ahead to attain dependable algorithms in safety-critical systems such as ships. Some of the outstanding challenges to be addressed are highlighted in the following:

1- The first and foremost, ML techniques demand a large amount of reliable data and computational time to be adequately trained. A certain degree of fidelity should also be maintained in these data for the algorithm to function properly. This demand result in some challenges as outlined below:

1-1 As a result of the high computational cost, in most cases the algorithms are only trained on a limited set of scenarios or dimensions, thereby their generalisability is impacted. Hence, the algorithm trained in a restricted environment may not be able to extrapolate its acquired knowledge to realistic conditions.

1-2 Due to the safety-critical nature of this application, planning cannot be easily solved based on trial-and-error learning, particularly in realistic scenarios. An extensive high-fidelity simulation analysis is therefore required prior to any experimentation to instil confidence in the algorithm. The transition between simulation to practise remains a major challenge (Pina et al., 2021). In conventional algorithms, there are some tuning knobs to tweak the performance in real-world operations. However, the behaviour and tuning of the trained algorithms in real environments would be challenging especially if the simulation environment is not perfect. Therefore, specific consideration for "simulation to real" should be foreseen (Pina et al., 2021).

2- Based on 1-1, due to the existing learning and computational challenges, most of the developed approaches are not tested in a comprehensive situation. To circumvent this, adopting proper Verification and Validation (V&V) procedures, e.g., testing in Monte Carlo runs and state of the art simulators such as digital twins, is recommended. A statistical analysis based on appropriate performance indicators should be conducted to assess the functionality of developed ML approaches from various aspects of view, such as feasibility and covering the mission planning areas presented in Section 3.

3- To address safety constraints in practice and prove the resilient behaviour of developed algorithms, one suggestion is to use ML algorithms to develop a captain-assistive system (Du et al., 2020) advising the captain on potential feasible paths. Further training of those algorithms and examination of their performance under controlled conditions could potentially lead to full autonomy in the future.

4- Based on this survey, model-free DRL appears to be the dominant category of approaches in planning and collision avoidance of vessels. A possible explanation is the affinity of reinforcement learning with control. Nevertheless, reward function design remains a grand challenge in DRLs. Utilising complicated rewards with several indices (as seen in some papers) may transform the DRL into supervised learning (Kendall et al., 2019). In this case, the definition of the reward and its index weights could become a non-trivial task.

5- This subject offers a variety of novel topics to explore, such as the development of realistic simulation tools like digital twins (Almeaibed, 2021, Vasanthan, and Nguyen, 2021), defining proper testing procedures and edge cases for algorithm acceptance (Perera, 2020), automatic test scenario generation (Riedmaier, et al., 2020), and leveraging ML techniques in animal behaviour modelling for marine animal obstacle avoidance (Schoeman et al., 2020).

## 7. CONCLUSIONS

This article surveyed recent advances in ML application for ship collision avoidance and mission planning. An exponentially increasing trend of published research on relevant topics has been identified in this review. Based on the pivotal areas of mission planning of autonomous ships, the available research was classified into two groups. The first group presented ML algorithms with direct application to collision avoidance and local planning. The second category included techniques that could be utilised in other mission planning applications such as global planning or risk assessment. Although various ML algorithms are adopted in the second category, it is found that DRL techniques (such as DQN, DDPG, PPO, etc.) are often used for local planning and collision avoidance. The choice may have been influenced by the analogy between reinforcement learning and the inner planning and control loops in autonomous vehicles. Last but not least, the main achievements, challenges, and future directions in this field were outlined. Within the next few years, it is likely that ML techniques will be implemented on autonomous ships. The challenge will be realizing those approaches as commercial products that are reliable and safe.

## ACKNOWLEDGEMENT

The authors would like to thank the UK Research and Innovation for funding this project which is part of the Belfast Maritime Consortium under the Strength in Places Funding programme.

## REFERENCES

Abebe, M., Noh, Y., Seo, C., Kim, D., & Lee, I. (2021). Developing a Ship Collision Risk Index estimation model based on Dempster-Shafer theory. Applied Ocean Research, 113, 102735.

Abebe, M., Noh, Y., Kang, Y. J., Seo, C., Kim, D., & Seo, J. (2022). Ship trajectory planning for collision avoidance using hybrid ARIMA-LSTM models. Ocean Engineering, 256, 111527.

Almeaibed, S., Al-Rubaye, S., Tsourdos, A., & Avdelidis, N. P. (2021). Digital twin analysis to promote safety and security in autonomous vehicles. IEEE Communications Standards Magazine, 5(1), 40-46.

Amendola, J., Miura, L. S., Costa, A. H. R., Cozman, F. G., & Tannuri, E. A. (2020). Navigation in Restricted Channels Under Environmental Conditions: Fast-Time Simulation by Asynchronous Deep Reinforcement Learning. IEEE Access, 8, 149199-149213.

Aradi, S. (2020). Survey of deep reinforcement learning for motion planning of autonomous vehicles. IEEE Transactions on Intelligent Transportation Systems.

Artemis (2020). "Artemis Technologies to build zero emissions ferries following £60M funding", https://www.artemistechnologies.co.uk/artemis-technologies-to-build-zero-emissions-ferries-following-60m-funding/, visited in March 2022.

Campbell, S., Naeem, W., & Irwin, G. W. (2012). A review on improving the autonomy of unmanned surface vehicles through intelligent collision avoidance manoeuvres. Annual Reviews in Control, 36(2), 267-283.

Cetus, (2022), "Uncrewed Surface Vessel (USV) Cetus for marine data gathering and systems development", https://www.plymouth.ac.uk/research/esif-funded-projects/usv-cetus, visited in March 2022.

Chen, P., Huang, Y., Mou, J., & Van Gelder, P. H. A. J. M. (2019a). Probabilistic risk analysis for ship-ship collision: State-of-the-art. Safety science, 117, 108-122.

Chen, C., Chen, X. Q., Ma, F., Zeng, X. J., & Wang, J. (2019b). A knowledge-free path planning approach for smart ships based on reinforcement learning. Ocean Engineering, 189, 106299.

Chen, C., Ma, F., Xu, X., Chen, Y., & Wang, J. (2021a). A novel ship collision avoidance awareness approach for cooperating ships using multi-agent deep reinforcement learning. Journal of Marine Science and Engineering, 9(10), 1056.

Chen, X., Liu, Y., Achuthan, K., Zhang, X., & Chen, J. (2021b). A semi-supervised deep learning model for ship encounter situation classification. Ocean Engineering, 239, 109824.

Cheng, Y., & Zhang, W. (2018). Concise deep reinforcement learning obstacle avoidance for underactuated unmanned marine vessels. Neurocomputing, 272, 63-73.

Chun, D. H., Roh, M. I., Lee, H. W., Ha, J., & Yu, D. (2021). Deep reinforcement learning-based collision avoidance for an autonomous ship. Ocean Engineering, 234, 109216.

Cui, Y., Osaki, S., & Matsubara, T. (2019). Reinforcement learning boat autopilot: a sample-efficient and model predictive control based approach. In 2019 IEEE/RSJ International Conference on Intelligent Robots and Systems (IROS) (pp. 2868-2875). IEEE.

Cui, Y., Osaki, S., & Matsubara, T. (2021). Autonomous boat driving system using sample-efficient model predictive control-based reinforcement learning approach. Journal of Field Robotics, 38(3), 331-354.

Du, L., Banda, O. A. V., Goerlandt, F., Huang, Y., & Kujala, P. (2020). A COLREG-compliant ship collision alert system for stand-on vessels. Ocean Engineering, 218, 107866.

Du, L., Banda, O. A. V., Huang, Y., Goerlandt, F., Kujala, P., & Zhang, W. (2021). An empirical ship domain based on evasive maneuver and perceived collision risk. Reliability Engineering & System Safety, 213, 107752.

Du, B., Lin, B., Zhang, C., Dong, B., & Zhang, W. (2022). Safe deep reinforcement learning-based adaptive control for USV interception mission. Ocean Engineering, 246, 110477.

Fan, Y., Sun, Z., & Wang, G. (2022). A Novel Reinforcement Learning Collision Avoidance Algorithm for USVs Based on Maneuvering Characteristics and COLREGs. Sensors, 22(6), 2099.

Fossen, T. I. (2011). Handbook of marine craft hydrodynamics and motion control. John Wiley & Sons.

Fraga-Lamas, P., Ramos, L., Mondéjar-Guerra, V., & Fernández-Caramés, T. M. (2019). A review on IoT deep learning UAV systems for autonomous obstacle detection and collision avoidance. Remote Sensing, 11(18), 2144.

Gao, M., & Shi, G. Y. (2020a). Ship collision avoidance anthropomorphic decision-making for structured learning based on AIS with Seq-CGAN. Ocean Engineering, 217, 107922.

Gao, M., & Shi, G. Y. (2020b). Ship-Collision Avoidance Decision-Making Learning of Unmanned Surface Vehicles with Automatic Identification System Data Based on Encoder—Decoder Automatic-Response Neural Networks. Journal of Marine Science and Engineering, 8(10), 754.

Gao, M., Kang, Z., Zhang, A., Liu, J., & Zhao, F. (2022). MASS autonomous navigation system based on AIS big data with dueling deep Q networks prioritized replay reinforcement learning. Ocean Engineering, 249, 110834.

Gjærum, V. B., Rørvik, E. L. H., & Lekkas, A. M. (2021a). Approximating a deep reinforcement learning docking agent using linear model trees. In 2021 European Control Conference (ECC) (pp. 1465-1471). IEEE.

Gjærum, V. B., Strümke, I., Alsos, O. A., & Lekkas, A. M. (2021b). Explaining a Deep Reinforcement Learning Docking Agent Using Linear Model Trees with User Adapted Visualization. Journal of Marine Science and Engineering, 9(11), 1178.

Goodfellow, I., Bengio, Y., & Courville, A. (2016). Deep learning. MIT press.

Gonzalez-Garcia, A., Castañeda, H., & Garrido, L. (2020). USV Path-Following Control Based On Deep Reinforcement Learning and Adaptive Control. In Global Oceans 2020: Singapore–US Gulf Coast (pp. 1-7). IEEE.

Guo, S., Zhang, X., Zheng, Y., & Du, Y. (2020). An autonomous path planning model for unmanned ships based on deep reinforcement learning. Sensors, 20(2), 426.

Hadi, B., Khosravi, A., & Sarhadi, P. (2021). A review of the path planning and formation control for multiple autonomous underwater vehicles. Journal of Intelligent & Robotic Systems, 101(4), 1-26.

Haydari, A., & Yilmaz, Y. (2020). Deep reinforcement learning for intelligent transportation systems: A survey. IEEE Transactions on Intelligent Transportation Systems.

Heiberg, A., Larsen, T. N., Meyer, E., Rasheed, A., San, O., & Varagnolo, D. (2022). Risk-based implementation of COLREGs for autonomous surface vehicles using deep reinforcement learning. Neural Networks, 152, 17-33.

Huang, Y., Chen, L., Chen, P., Negenborn, R. R., & Van Gelder, P. H. A. J. M. (2020a). Ship collision avoidance methods: State-of-the-art. Safety science, 121, 451-473.

Huang, Y., & Van Gelder, P. H. A. J. M. (2020b). Collision risk measure for triggering evasive actions of maritime autonomous surface ships. Safety science, 127, 104708.

Ibarz, J., Tan, J., Finn, C., Kalakrishnan, M., Pastor, P., & Levine, S. (2021). How to train your robot with deep reinforcement learning: lessons we have learned. The International Journal of Robotics Research, 40(4-5), 698-721.

IMO (1972). Convention on the International Regulations for Preventing Collisions at Sea. Available: https://www.imo.org/en/About/Conventions/Pages/COLREG.aspx.

Ivanov, Y. S., Zhiganov, S. V., & Ivanova, T. I. (2021). Intelligent deep neuro-fuzzy system of abnormal situation recognition for transport systems. In Current Problems and Ways of Industry Development: Equipment and Technologies (pp. 224-233). Springer, Cham.

Jeong-Seok, L. E. E., Hyeong-Tak, L. E. E., & Ik-Soon, C. H. O. (2022). Developing a Maritime Traffic Route Framework Based on Statistical Density Analysis from AIS Data Using a Clustering Algorithm. IEEE Access.

Kendall, A., Hawke, J., Janz, D., Mazur, P., Reda, D., Allen, J. M., & Shah, A. (2019). Learning to drive in a day. In 2019 International Conference on Robotics and Automation (ICRA) (pp. 8248-8254).

Kim, K. I., & Lee, K. M. (2018). Deep learning-based caution area traffic prediction with automatic identification system sensor data. Sensors, 18(9), 3172.

Kiran, B. R., Sobh, I., Talpaert, V., Mannion, P., Al Sallab, A. A., Yogamani, S., & Pérez, P. (2021). Deep reinforcement learning for autonomous driving: A survey. IEEE Transactions on Intelligent Transportation Systems.

Kroemer, O., Niekum, S., & Konidaris, G. D. (2021). A review of robot learning for manipulation: Challenges, representations, and algorithms. Journal of machine learning research, 22(30).

Kuutti, S., Bowden, R., Jin, Y., Barber, P., & Fallah, S. (2021). A survey of deep learning applications to autonomous vehicle control. IEEE Transactions on Intelligent Transportation Systems, 22(2), 712-733.

L3HARRIS (2021), "L3HARRIS technologies to design long-endurance autonomous surface ship concept for us defense advanced research projects agency", https://www.l3harris.com/newsroom/press-release/2021/03/l3harris-technologies-design-long-endurance-autonomous-surface-ship, visited in March 2022.

Lei, P. R., Yu, P. R., & Peng, W. C. (2019). A framework for maritime anti-collision pattern discovery from AIS network. In 2019 20th Asia-Pacific Network Operations and Management Symposium (APNOMS) (pp. 1-4). IEEE.

Lei, P. R., Yu, P. R., & Peng, W. C. (2021). Learning for Prediction of Maritime Collision Avoidance Behavior from AIS Network. In 2021 22nd Asia-Pacific Network Operations and Management Symposium (APNOMS) (pp. 222-225). IEEE.

Li, Y. (2017). Deep reinforcement learning: An overview. arXiv preprint arXiv:1701.07274.

Li, M., Mou, J., Chen, L., Huang, Y., & Chen, P. (2021a). Comparison between the collision avoidance decision-making in theoretical research and navigation practices. Ocean Engineering, 228, 108881.

Li, L., Wu, D., Huang, Y., & Yuan, Z. M. (2021b). A path planning strategy unified with a COLREGS collision avoidance function based on deep reinforcement learning and artificial potential field. Applied Ocean Research, 113, 102759.

Liu, X., & Jin, Y. (2020). Reinforcement learning-based collision avoidance: impact of reward function and knowledge transfer. AI EDAM, 34(2), 207-222.

Luis, S. Y., Reina, D. G., & Marín, S. L. T. (2021). A multiagent deep reinforcement learning approach for path planning in autonomous surface vehicles: The Ypacaraí lake patrolling case. IEEE Access, 9, 17084-17099.

Ma, Y., Wang, Z., Yang, H., & Yang, L. (2020). Artificial intelligence applications in the development of autonomous vehicles: a survey. IEEE/CAA Journal of Automatica Sinica, 7(2), 315-329.

Martinsen, A. B., Lekkas, A. M., & Gros, S. (2022). Reinforcement learning-based NMPC for tracking control of ASVs: Theory and experiments. Control Engineering Practice, 120, 105024.

Mayflower (2022), "The Uncharted: Autonomous Ship Project No captain. No crew. No problem", https://www.ibm.com/industries/federal/autonomous-ship, visited in March 2022.

MAXCMAS, (2018). "MAXCMAS success suggests COLREGs remain relevant for autonomous ships", https://www.rolls-royce.com/media/press-releases/2018/21-03-2018-maxcmas-success-suggests-colregs-remain-relevant-for-autonomous-ships.aspx, visited in March 2022.

Meyer, E., Robinson, H., Rasheed, A., & San, O. (2020a). Taming an autonomous surface vehicle for path following and collision avoidance using deep reinforcement learning. IEEE Access, 8, 41466-41481.

Meyer, E., Heiberg, A., Rasheed, A., & San, O. (2020b). COLREG-compliant collision avoidance for unmanned surface vehicle using deep reinforcement learning. IEEE Access, 8, 165344-165364.

Murray, B., & Perera, L. P. (2021a). An AIS-based deep learning framework for regional ship behavior prediction. Reliability Engineering & System Safety, 215, 107819.

Murray, B., & Perera, L. P. (2021b). Deep representation learning-based vessel trajectory clustering for situation awareness in ship navigation. In Developments in Maritime Technology and Engineering (pp. 157-165). CRC Press.

Murray, B., & Perera, L. P. (2021c). Proactive Collision Avoidance for Autonomous Ships: Leveraging Machine Learning to Emulate Situation Awareness. IFAC-PapersOnLine, 54(16), 16-23.

Namgung, H., & Kim, J. S. (2021a). Collision risk inference system for maritime autonomous surface ships using COLREGs rules compliant collision avoidance. IEEE Access, 9, 7823-7835.

Namgung, H., & Kim, J. S. (2021b). Regional Collision Risk Prediction System at a Collision Area Considering Spatial Pattern. Journal of Marine Science and Engineering, 9(12), 1365.

Ozturk, U., Birbil, S. I., & Cicek, K. (2019). Evaluating navigational risk of port approach manoeuvrings with expert assessments and machine learning. Ocean Engineering, 192, 106558.

Ozturk, Ü., Akdağ, M., & Ayabakan, T. (2022). A review of path planning algorithms in maritime autonomous surface ships: Navigation safety perspective. Ocean Engineering, 251, 111010.

Park, J., & Jeong, J. S. (2021). An Estimation of Ship Collision Risk Based on Relevance Vector Machine. Journal of Marine Science and Engineering, 9(5), 538.

Perera, L. P. (2020). Deep learning toward autonomous ship navigation and possible COLREGs failures. Journal of Offshore Mechanics and Arctic Engineering, 142(3).

Pietrzykowski, Z., & Wielgosz, M. (2021). Effective ship domain–Impact of ship size and speed. Ocean Engineering, 219, 108423.

Pina, R., Tibebu, H., Hook, J., De Silva, V., & Kondoz, A. (2021). Overcoming Challenges of Applying Reinforcement Learning for Intelligent Vehicle Control. Sensors, 21(23), 7829.

Riedmaier, S., Ponn, T., Ludwig, D., Schick, B., & Diermeyer, F. (2020). Survey on scenario-based safety assessment of automated vehicles. IEEE access, 8, 87456-87477.

Rawson, A., Brito, M., Sabeur, Z., & Tran-Thanh, L. (2021). A machine learning approach for monitoring ship safety in extreme weather events. Safety science, 141, 105336.

Rawson, A., & Brito, M. (2021). Developing contextually aware ship domains using machine learning. The Journal of Navigation, 74(3), 515-532.

Sarhadi, P., Naeem, W., & Athanasopoulos, N. (2022, April). An Integrated Risk Assessment and Collision Avoidance Methodology for An Autonomous Catamaran with Fuzzy Weighting Functions. In 2022 UKACC 13th International Conference on Control, Plymouth, UK, (pp. 228-234). IEEE.

Sawada, R., Sato, K., & Majima, T. (2021). Automatic ship collision avoidance using deep reinforcement learning with LSTM in continuous action spaces. Journal of Marine Science and Technology, 26(2), 509-524.

Schoeman, R. P., Patterson-Abrolat, C., & Plön, S. (2020). A global review of vessel collisions with marine animals. Frontiers in Marine Science, 7, 292.

Shen, H., Hashimoto, H., Matsuda, A., Taniguchi, Y., Terada, D., & Guo, C. (2019). Automatic collision avoidance of multiple ships based on deep Q-learning. Applied Ocean Research, 86, 268-288.

Shirobokov, M., Trofimov, S., & Ovchinnikov, M. (2021). Survey of machine learning techniques in spacecraft control design. Acta Astronautica, 186, 87-97.

Shi, J. H., & Liu, Z. J. (2020). Deep Learning in Unmanned Surface Vehicles Collision-Avoidance Pattern Based on AIS Big Data with Double GRU-RNN. Journal of Marine Science and Engineering, 8(9), 682.

Sun, H., Zhang, W., Runxiang, Y. U., & Zhang, Y. (2021). Motion planning for mobile Robots–focusing on deep reinforcement learning: A systematic Review. IEEE Access.

Sutton, R. S., & Barto, A. G. (2018). Reinforcement learning: An introduction. MIT press.

Tam, C., Bucknall, R. and Greig, A., (2009). Review of collision avoidance and path planning methods for ships in close range encounters. The Journal of Navigation, 62(3), pp.455-476.

Vasanthan, C., & Nguyen, D. T. (2021). Combining Supervised Learning and Digital Twin for Autonomous Path-planning. IFAC-PapersOnLine, 54(16), 7-15.

Vagale, A., Oucheikh, R., Bye, R. T., Osen, O. L., & Fossen, T.. (2021a). Path planning and collision avoidance for autonomous surface vehicles I: a review. Journal of Marine Science and Technology, 1-15.

Vagale, A., Oucheikh, R., Bye, R. T., Osen, O. L., & Fossen, T.. (2021b). Path planning and collision avoidance for autonomous surface vehicles II: a comparative study of algorithms. Journal of Marine Science and Technology, 1-17.

Wang, S., Ma, F., Yan, X., Wu, P., & Liu, Y. (2021). Adaptive and extendable control of unmanned surface vehicle formations using distributed deep reinforcement learning. Applied Ocean Research, 110, 102590.

Wang, W., Luo, X., Li, Y., & Xie, S. (2021). Unmanned surface vessel obstacle avoidance with prior knowledge-based reward shaping. Concurrency and Computation: Practice and Experience, 33(9), e6110.

Woo, J., & Kim, N. (2020). Collision avoidance for an unmanned surface vehicle using deep reinforcement learning. Ocean Engineering, 199, 107001.

Wu, X., Chen, H., Chen, C., Zhong, M., Xie, S., Guo, Y., & Fujita, H. (2020). The autonomous navigation and obstacle avoidance for USVs with ANOA deep reinforcement learning method. Knowledge-Based Systems, 196, 105201.

Xie, S., Garofano, V., Chu, X., & Negenborn, R. R. (2019). Model predictive ship collision avoidance based on Q-learning beetle swarm antenna search and neural networks. Ocean Engineering, 193, 106609.

Xie, S., Chu, X., Zheng, M., & Liu, C. (2020). A composite learning method for multi-ship collision avoidance based on reinforcement learning and inverse control. Neurocomputing, 411, 375-392.

Xu, H., Wang, N., Zhao, H., & Zheng, Z. (2019). Deep reinforcement learning-based path planning of underactuated surface vessels. Cyber-Physical Systems, 5(1), 1-17.

Xu, X., Lu, Y., Liu, X., & Zhang, W. (2020). Intelligent collision avoidance algorithms for USVs via deep reinforcement learning under COLREGs. Ocean Engineering, 217, 107704.

Xu, X., Cai, P., Ahmed, Z., Yellapu, V. S., & Zhang, W. (2022a). Path planning and dynamic collision avoidance algorithm under COLREGs via deep reinforcement learning. Neurocomputing, 468, 181-197.

Xu, X., Lu, Y., Liu, G., Cai, P., & Zhang, W. (2022b). COLREGs-abiding hybrid collision avoidance algorithm based on deep reinforcement learning for USVs. Ocean Engineering, 247, 110749.

Yara, (2021). "Yara to start operating the world's first fully emission-free container ship", https://www.yara.com/corporate-releases/yara-to-start-operating-the-worlds-first-fully-emission-free-container-ship/, visited in March 2022.

Zhai, P., Zhang, Y., & Shaobo, W. (2022). Intelligent Ship Collision Avoidance Algorithm Based on DDQN with Prioritized Experience Replay under COLREGs. Journal of Marine Science and Engineering, 10(5), 585.

Zhao, L., & Roh, M. I. (2019). COLREGs-compliant multiship collision avoidance based on deep reinforcement learning. Ocean Engineering, 191, 106436.

Zhao, L., Roh, M. I., & Lee, S. J. (2019). Control method for path following and collision avoidance of autonomous ship based on deep reinforcement learning. Journal of Marine Science and Technology, 27(4), 1.

Zhao, J., Lu, J., Chen, X., Yan, Z., Yan, Y., & Sun, Y. (2022). High-fidelity data supported ship trajectory prediction via an ensemble machine learning framework. Physica A: Statistical Mechanics and Its Applications, 586, 126470.

Zhang, X., Wang, C., Jiang, L., An, L., & Yang, R. (2021a). Collision-avoidance navigation systems for Maritime Autonomous Surface Ships: A state of the art survey. Ocean Engineering, 235, 109380.

Zhang, Q., Pan, W., & Reppa, V. (2021b). Model-reference reinforcement learning for collision-free tracking control of autonomous surface vehicles. IEEE Transactions on Intelligent Transportation Systems.

Zhou, X., Wu, P., Zhang, H., Guo, W., & Liu, Y. (2019). Learn to navigate: cooperative path planning for unmanned surface vehicles using deep reinforcement learning. IEEE Access, 7, 165262-165278.

Zhou, C., Wang, Y., Wang, L., & He, H. (2022). Obstacle avoidance strategy for an autonomous surface vessel based on modified deep deterministic policy gradient. Ocean Engineering, 243, 110166.